%% file: main.tex
\documentclass[10pt,twocolumn,letterpaper]{article}

\usepackage{wacv}              

\usepackage{graphicx}%
\usepackage{multirow}%
\usepackage{amsmath,amssymb,amsfonts}%
\usepackage{amsthm}%
\usepackage[accsupp]{axessibility}
\usepackage{mathrsfs}%
\usepackage[title]{appendix}%
\usepackage{xcolor}%
\usepackage{textcomp}%
\usepackage{manyfoot}%
\usepackage{booktabs}%
\usepackage{siunitx}%
\usepackage{algorithm}%
\usepackage{algorithmicx}%
\usepackage{algpseudocode}%
\usepackage{listings}%
\usepackage{xspace}%
\usepackage[utf8]{inputenc}%
\usepackage{microtype}
\usepackage{times}
\usepackage{epsfig}
\usepackage{lipsum}
\usepackage{subcaption}
\usepackage{wrapfig}
\usepackage{enumitem}
\usepackage{color}
\usepackage{colortbl}
\usepackage{sidecap}

\usepackage[pagebackref,breaklinks,colorlinks]{hyperref}
\usepackage[capitalize]{cleveref}
\crefname{section}{sec.}{secs.}
\Crefname{section}{Sec.}{Secs.}
\crefname{table}{tab.}{tabs.}
\Crefname{table}{Tab.}{Tabs.}
\crefname{figure}{fig.}{figs.}
\Crefname{figure}{Fig.}{Figs.}
\crefname{equation}{eq.}{eqs.}
\Crefname{equation}{Eq.}{Eqs.}

\def\wacvPaperID{890} 
\def\confName{WACV}
\def\confYear{2025}

\begin{document}

\title{DarSwin-Unet: Distortion Aware Architecture}

\author{Akshaya Athwale*{$^1$}, Ichrak Shili*{$^1$}, Émile Bergeron{$^1$}, Ola Ahmad{$^2$}, Jean-François Lalonde{$^1$} \\
{\small{$^1$}Université Laval \quad {$^2$}Thales CortAIx Labs Canada}
}

\maketitle

\newcommand{\todo}[1]{\textcolor{red}{[\textbf{TODO}: #1]}}
\newcommand{\ola}[1]{\textcolor{blue}{[\textbf{ola}: #1]}}
\newcommand{\ak}[1]{\textcolor{cyan}{#1}}
\newcommand{\ichrak}[1]{\textcolor{orange}{#1}}
\newcommand{\emile}[1]{\textcolor{green}{#1}}

\definecolor{best}{RGB}{255, 220, 200}
\definecolor{second}{RGB}{255, 255, 200}

\newcommand{\thename}{DarSwin-Unet\xspace}
\newcommand{\ogname}{DarSwin-Unet\xspace}

\input{sections/abstract.tex}
\input{sections/introduction.tex}
\input{sections/related_works.tex}

\input{sections/background.tex}
\input{sections/method.tex}
\input{sections/experiments.tex}

\input{sections/discussion.tex}
\input{sections/acknowledgment}

{\small
\bibliographystyle{ieee_fullname}
\bibliography{allrefs,egbib}
}

\end{document}


\title{DarSwin-Unet: Distortion Aware Architecture-Supplementary Material}


\twocolumn[{%
\renewcommand\twocolumn[1][]{#1}%
\maketitle
\begin{center}
    \centering
    \captionsetup{type=figure}
    \includegraphics[width=0.8\linewidth]{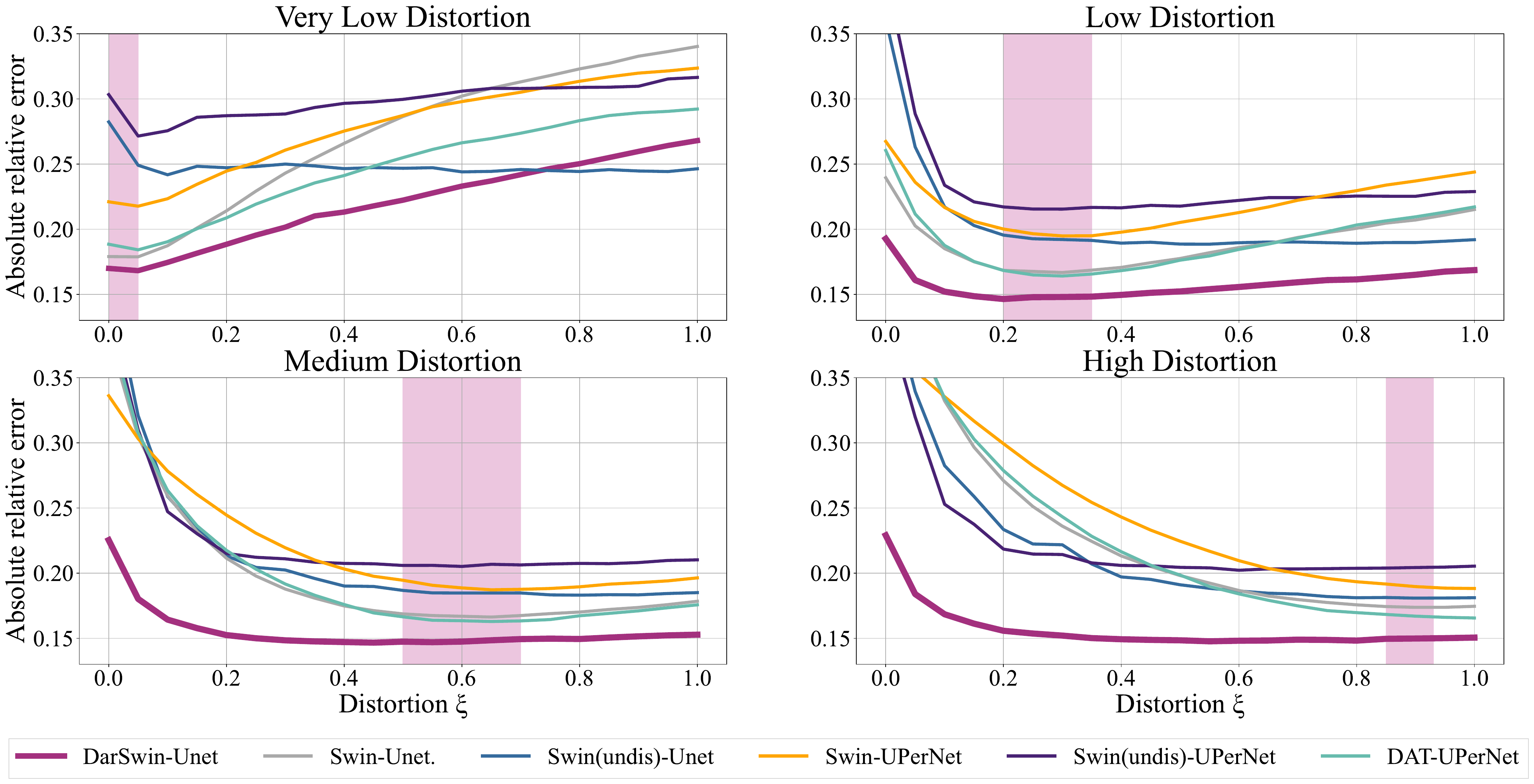}
    \captionof{figure}{Absolute relative error (lower better) in depth estimation as a function of test distortion for the six baselines: \thename, Swin-Unet \cite{cao2021swinunet}, Swin(undis)-Unet, Swin-UPerNet \cite{liu2021swin, xiao2018unified}, Swin(undis)-UPerNet and DAT-UPerNet \cite{xia2022vision, xiao2018unified}. All methods are trained on a restricted set of lens distortion curves (indicated by the pink shaded regions): (a) very low, (b) low, (c) medium, and (d) high distortion. We study the generalization abilities of each model by testing across all $\xi \in [0, 1]$. The squared relative error follows the same curves as the absolute relative error.}
\end{center}%
}]

\def\thesection{\Alph{section}}

\section{Depth estimation}
\label{sec:depth_supp}

\subsection{Evaluation metrics}
These detailed explanation of the evaluation metrics as shown in main text are defined as follows: 
\begin{itemize}
    \item Absolute relative error $=
      \frac{1}{|D|} \sum_{d \in D}^{} \frac{|d^* - d |}{d^*}$
    \item RMSE $=
    \sqrt{\frac{1}{|D|} \sum_{d \in D}^{} ||d^* - d ||^2}$
    \item Square relative error $=
      \frac{1}{|D|} \sum_{d \in D}^{} \frac{||d^* - d ||^2}{d^*}$
    \item log-RMSE $=
    \sqrt{\frac{1}{|D|} \sum_{d \in D}^{} || \log d^* - \log d ||^2}$
    \item \textbf{$\delta_t$} $= \frac{1}{|D|} |\{ d\in D | \max(\frac{d^*}{d},\frac{d}{d^*}) \leq 1.25^t\}|, \quad t \in \{1,2,3\}$
\end{itemize}
with $D$, $d^*$ and $d$ are respectively the set of valid depths, the ground truth depth and the predicted depth. We show the results for each metric similar to Fig. 9 in the main text.

\begin{figure*}[!ht]
    \centering
    \includegraphics[width=1.0\linewidth]{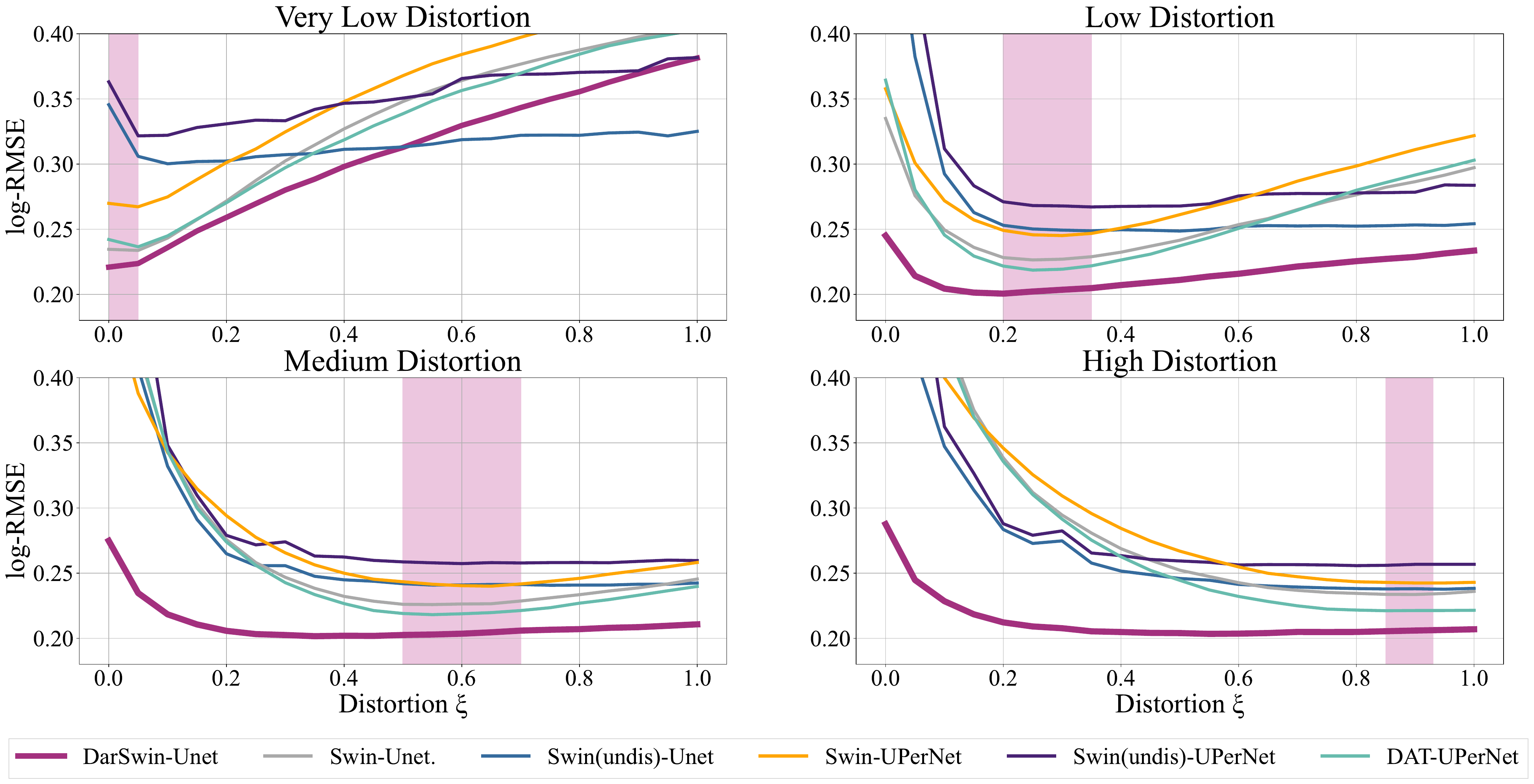}
    \caption[]{log-RMSE (lower better) in depth estimation as a function of test distortion for the six baselines: \thename, Swin-Unet \cite{cao2021swinunet}, Swin(undis)-Unet, Swin-UPerNet \cite{liu2021swin, xiao2018unified}, Swin(undis)-UPerNet and DAT-UPerNet \cite{xia2022vision, xiao2018unified} . All methods are trained on a restricted set of lens distortion curves (indicated by the pink shaded regions): (a) Very low, (b) low, (c) medium and (d) high distortion. We study the generalization abilities of each model by testing across all $\xi \in [0, 1]$. RMSE follows the same curves as log-RMSE.}
    \label{fig:results-depth-log-rmse}
\end{figure*}

\subsection{Proposed sampling function}

The goal is to identify a class of functions that is parameterized by a minimal number of parameters while still being capable of representing a wide variety of monotonic profiles between two interpolation points, $(0,0)$ and $(a,b)$. The initial approach involves using a power-law function, which is widely employed in engineering due to its simplicity and its ability to model relationships between unknown quantities with minimal parametrization.

\begin{align}
    p_n(\theta) = b\left(\frac{\theta}{a}\right)^n \,, \nonumber 
\end{align}

The function $p(0) = 0 \text{ and } p(a) = \text{FOV}$. This formulation generates convex curves $(n \geq 1)$ or concave curves $(n < 1)$, with a derivative of zero or a non-existent derivative (tangent to the y-axis) at the origin. The underlying idea is that if a curve exhibits cuspidal behavior at one end, it should also be capable of exhibiting such behavior at the other end. To achieve this symmetry, two reflections are applied to flip the function vertically and horizontally.

\begin{align}
    q_m(\theta) = 1 - \left( 1 - \frac{\theta}{a}\right)^m \,, \nonumber 
\end{align}

which also satisfy the interpolation conditions. This approach can generate both convex $(m \geq 1)$ and concave $(m < 1)$  curves. To combine these curves while ensuring the interpolation conditions remain satisfied, their convex combination is utilized.
\begin{align}
    g(\theta) = \lambda p_n(\theta) + (1-\lambda)q_m(\theta)\,,
\end{align}

for $\lambda \in [0, 1]$. If $m < 1$, $n > 1$ or $m > 1$, $n < 1$, the resulting curve is clearly monotonic increasing. In cases where both $m$ and $n$ are either greater than 1 or less than 1, the curve remains monotonic. This family of curves is parameterized by the three parameters $m, n, t$.


\subsection{Derivative of uniform camera model projection}
The Unified camera model~\cite{barreto2006unifying,mei2007single} as explained in the main text, \emph{bounded} parameter $\xi \in [0, 1]$\footnote{$\xi$ can be slightly greater than 1 for certain types of catadioptric cameras~\cite{ying2004consider} but this is ignored here.} projects the world point to the image as follows
\begin{equation}
r_d = \mathcal{P}(\theta) = \frac{f\cos\theta}{\xi + \sin{\theta}} \,,
\label{eqn:sphproj}
\end{equation}
where $r_d$ is the radial distance from the image center, $\theta$ the incident angle lens, $f$ the focal length and $\xi$ the distortion parameter.

For a fixed $\theta$ (field of view), to prove that the extremities of the derivatives with respect to $g(\theta)$ for this projection function lie at $\xi = 0$ and $\xi = 1$, we first need to calculate $\diff{r_d}{g(\theta)}$. To do so, let us first compute $\diff{r_d}{\theta}$,

\begin{align}
&\diff{r_d}{\theta} = \diff{}{\theta}(\frac{f\cos\theta}{\xi + \sin{\theta}}) \,,\\
&\diff{r_d}{\theta} = \frac{\diff{}{\theta}(f\cos(\theta))(\xi + \sin(\theta)) - (f\cos(\theta))\diff{}{\theta}(\xi + \sin(\theta))}{(\xi + \sin(\theta))^2} \,, \\
&\diff{r_d}{\theta} = \frac{-f(\xi\sin{\theta} + 1)}{(\xi + \sin(\theta))^2}\, .
\end{align}

To calculate $\diff{r_d}{g(\theta)}$, we can write $\theta = g^{-1}(g(\theta))$ and use chain rule : 

\begin{align}
&\diff{\theta}{g(\theta)} = \frac{1}{g^\prime(\theta)} \,, \\
&\diff{r_d}{g(\theta)} = \diff{r_d}{\theta}\diff{\theta}{g(\theta)}\,, \\
&\diff{r_d}{g(\theta)} = \frac{-f(\xi\sin{\theta} + 1)}{g^\prime(\theta)(\xi + \sin(\theta))^2}\,.
\nonumber 
\end{align}

To prove that the extremities of the derivative of the projection function occur at $\xi = 0 \text{ and } \xi = 1$, we need to prove that the derivative is monotonic, i.e. $\diff{}{\xi}(\diff{r_d}{g(\theta)}) > 0$, Fist we analysis this derivative, using the quotient rule, the derivative becomes: 

\begin{align*}
    &\diff{}{\xi}\left( \frac{-f(\xi \sin\theta + 1)}{g'(\theta)(\xi + \sin\theta)^2} \right) = \\
&\frac{-f\sin\theta(\xi + \sin\theta)^2 + 2f(\xi \sin\theta + 1)(\xi + \sin\theta)}{g'(\theta)(\xi + \sin\theta)^4}\,.
\end{align*}

The denominator is $g'(\theta)(\xi + \sin\theta)^4$, since $g(\theta)$ is monotonic $g'(\theta) > 0$ and $(\xi + \sin\theta)^4 > 0$.

The numerator is:

\begin{align}    
    &N(\xi) = -f\sin\theta(\xi + \sin\theta)^2 \\
    &+ 2f(\xi \sin\theta + 1)(\xi + \sin\theta) \,.\nonumber
\end{align}

Let us analyze this numerator, we want $N(\xi) > 0$ as well, 

\begin{align}
    &2f(\xi \sin\theta + 1)(\xi + \sin\theta) > f\sin\theta(\xi + \sin\theta)^2 \,, \\  \nonumber
    &2(\xi \sin\theta + 1) > \sin\theta(\xi + \sin\theta) \text{ since } ((\xi + \sin\theta) \neq 0) \,, \\ \nonumber
    &\xi\sin\theta -\sin^2\theta + 2 >0\,,\\ \nonumber
    & 2 > \sin\theta(\sin\theta - \xi) \,. \nonumber
\end{align}

Since $\theta = \text{FOV}/2$ it follows that $\theta \in [0, \pi] \text{, } \sin\theta \in [0,1]$ and $\xi \in [0, 1]$. Therefore, the maximum value of $\sin\theta(\sin\theta - \xi)$ occurs at $\xi = 0 \text{ and } \sin\theta = 1$.

Therefore, $\diff{}{\xi}(\diff{r_d}{g(\theta)}) > 0$, meaning the derivative $\diff{r_d}{g(\theta)}$ is monotonic with respect to $\xi$. Consequently, the maximum value of this derivative $\diff{r_d}{g(\theta)}$ occurs either at $\xi = 0$ or $\xi = 1$.

{\small
\bibliographystyle{ieee_fullname}
\bibliography{egbib_supp}
}

%% file: sections/abstract.tex
\begin{abstract}

Wide angle fisheye images are becoming increasingly common for perception tasks in applications such as robotics, security, and mobility (e.g. drones, avionics). However, current models often either ignore the distortions in wide angle images or are not suitable to perform pixel-level tasks. In this paper, we present an encoder-decoder model based on a radial transformer architecture that adapts to distortions in wide angle lenses by leveraging the physical characteristics defined by the radial distortion profile. In contrast to the original model, which only performs classification tasks, we introduce a U-Net architecture, \thename, designed for pixel level tasks. Furthermore, we propose a novel  strategy that minimizes sparsity when sampling the image for creating its input tokens. Our approach enhances the model capability to handle pixel-level tasks in wide angle fisheye images, making it more effective for real-world applications. Compared to other baselines, \thename achieves the best results across different datasets, with significant gains when trained on bounded levels of distortions (very low, low, medium, and high) and tested on all, including out-of-distribution distortions. We demonstrate its performance on depth estimation and show through extensive experiments that \thename can perform zero-shot adaptation to unseen distortions of different wide angle lenses. The code and models are publicly available at \href{https://lvsn.github.io/darswin-unet/}{https://lvsn.github.io/darswin-unet/}. 

\end{abstract}

%% file: sections/introduction.tex
\section{Introduction}

\footnote{*Authors contributed equally}
Many areas in computer vision, such as security \cite{kim2015fisheye}, augmented reality (AR) \cite{schmalstieg2017ar}, healthcare, and particularly autonomous vehicles \cite{deng2019restricted,yogamani2019woodscapes}, utilize wide angle lenses because they minimize costs by requiring fewer cameras to capture a $360^\circ$ scene due to their increased field of view.

However, this cost benefit comes with a drawback: images captured by wide angle lenses exhibit significant distortion because the projection model is no longer perspective. Straight lines in the real world appear curved in the image, and the geometry of objects changes as a function of their location in the image. The majority of CNN-based models have an implicit bias towards perspective images---indeed, the distortions in wide angle lenses break the translational equivariance of CNNs, limiting their applicability. The diversity in wide angle lens distortions further exacerbates this problem: a method trained on a specific lens distortion does not generalize well when evaluated on another lens with different distortion---one must therefore repeat the entire data collection, training procedure, etc. on such a new lens. 

 
One popular strategy to improve generalization when tested on another lens is canceling the distortion effect by warping the input image back to the perspective projection model. A wide array of such methods, ranging from classical~\cite{brousseau2019calibration,ramalingan2017unifying,zhang2015line,melo2013unsupervised,kannala2006generic} to deep learning~\cite{yin2018fisheyerectnet,xue2019learning}, have been proposed to train and test on the undistorted image. Unfortunately, canceling the effect of the distortion of wide angle images tends to create severely stretched images. It can also restrict the maximum field of view since, in the limit, a point at $90^\circ$ azimuth projects at infinity, but reducing the maximum field of view defeats the purpose of using a wide angle lens in the first place. Other projections are also possible (e.g., cylindrical~\cite{plaut2021fisheye}, or piecewise linear~\cite{yogamani2019woodscapes}), but these also tend to create unwanted distortions or suffer from resolution loss. Some methods like \cite{playout2021adaptable,ahmad2022fisheyehdk} use deformable convolutions~\cite{dai2017deformable,zhu2019deformable} to reason on wide angle image without undistorting them. Here, convolution kernels adapt to the lens distortion of the given image during training. However, these methods tend to overfit to the wide angle lens distortion present at training; hence, they cannot generalize over unseen lens distortion at test time. Transformer-based architectures~\cite{healswin,athwale2023darswin} are also used to reason directly on the wide angle image, even evaluating the generalization performance to other distortions in DarSwin~\cite{athwale2023darswin}. However, DarSwin was only demonstrated for classification, and HealSwin~\cite{healswin} could not adapt to other lenses at test time. To this date, it is not clear whether distortion-aware architectures can be trained for pixel-level tasks with zero-shot generalization to other distortion lenses at test time without fine-tuning.


In this work, we present a robust solution to that very problem. In particular we present an encoder-decoder architecture named \thename, which leverages DarSwin \cite{athwale2023darswin} as the encoder. While this strategy is effective, we observe that the image sampling pattern proposed in \cite{athwale2023darswin} creates sparsity issues which negatively affect the performance of pixel-level tasks such as depth estimation. 
To address this issue, we propose a novel pixel sampling method that mitigates the aforementioned sparsity problem, thereby significantly improving depth estimation performance. 
We present experiments on the depth estimation task, which show that \thename is much more robust to changes in lens distortions at test time than all of the compared baselines, including Swin-Unet~\cite{cao2021swinunet} and Swin-UPerNet~\cite{liu2021swin,xiong2019upsnet} trained on both distorted and undistorted images, and DAT-UPerNet~\cite{xia2022vision,xiong2019upsnet} trained on distorted images. 

In short, we make the following key contributions:
\begin{itemize}[nosep]
\item a novel encoder-decoder distortion-aware architecture, named \thename, suitable for pixel-level tasks, which adapts to the distortion in wide angle images in a zero-shot manner at test time, without fine-tuning; 
\item a new pixel sampling scheme which limits the sample sparsity problem when dealing with images of drastically different distortions;
\item extensive experiments on depth estimation showing the superiority of \thename at adapting to novel distortion profiles at test time.
\end{itemize}

%% file: sections/related_works.tex
\section{Related work}
\label{sec:related}

\paragraph*{Distortion correction}

Wide angle cameras are increasingly used in various computer vision applications, including visual perception \cite{kumar2021omnidet} and autonomous vehicle cameras \cite{yogamani2019woodscapes, kumar2022surround, liao2022kitti}. However, their adoption has been relatively recent \cite{Rashed_2021_WACV,inproceedings,plaut2021fisheye, Ye2020UniversalSS, yogamani2019woodscapes} due to the distortion present in their images. Earlier methods primarily focused on correcting this distortion \cite{article1, xue2019learning, DDM, yin2018fisheyerectnet, zhang2015line, liao2021ordinal, 9980359, yang2021progressively, feng2023simfir, yang2023dual}. Some approaches \cite{DDM, yang2022fishformer, liao2021ordinal} use distortion parameters to assess distortion density per pixel and subsequently correct it. However, such correction processes can introduce artifacts like stretching \cite{yogamani2019woodscapes}, leading to performance degradation.

In contrast, our approach builds upon DarSwin \cite{athwale2023darswin}, which directly utilizes distortion parameters to reason about wide angle images, avoiding the pitfalls associated with traditional distortion correction methods. This approach is crucial as it addresses distortions inherent in wide angle lenses and aligns with the needs of modern computer vision tasks.

\paragraph*{Convolution-based approaches.} 
CNNs \cite{krizhevsky2012classif, VGG, googlenet} are highly effective for processing images with no distortion due to their inherent bias towards natural image characteristics, such as translational equivariance \cite{GDL}. Methods like \cite{8500456, inproceedings, Rashed_2021_WACV} try to adapt CNNs on fisheye images for tasks such as object detection. However, the distortion caused by wide angle images breaks this symmetry, which reduces the performance of CNNs. Methods like \cite{yan2022fisheyedistill, omni, syndistnet, kumar2021svdistnet} use self-supervised learning combined with techniques like distillation or multi-task learning to have a better understanding of distortion. Deformable convolutions \cite{dai2017deformable, zhu2019deformable} offer flexibility by learning kernel deformations, though at a higher computational cost. Recent studies \cite{ahmad2022fisheyehdk, playout2021adaptable, deng2019restricted, Wei_2023_ICCV} use deformable CNNs to handle fisheye distortion. In contrast, Our network builds on DarSwin \cite{athwale2023darswin}, which leverages attention using a lens distortion profile instead of convolutions. However, unlike DarSwin’s encoder-only design, our network, \thename, introduces an encoder-decoder architecture.

\paragraph*{Hybrid-network based approaches.}
Some methods leverage properties from both self-attention and convolutions and build hybrid networks. Methods like \cite{Lee_2023_ICCV} use hybrid networks and try to leverage the geometric property of fisheye images (i.e., the orthogonal placement of objects) and propose a new representation of fisheye road scenes, invariant to the camera viewing direction. Shi et al.~\cite{shi2023fishdreamer} leverages the radial nature of distortion by including polar cross attention for inpainting, but unlike \thename, they do not use the lens information in their network. Similar to our method, \cite{kumar2021svdistnet, omni, kumar2020fisheyedistancenet, kumar2023unrectdepthnet} propose a camera-aware depth estimation network to handle the severe distortion of fisheye cameras: \cite{omni} encode the camera intrinsic parameters as a tensor; and \cite{kumar2023unrectdepthnet} propose a self-supervised depth estimation method which relies on the lens distortion parameter for forward and back-projection functions. Both these methods use distortion parameters as a part of the input or training process, but they rely on convolutions whose generalization capabilities are limited due to the translational invariance assumption being broken in wide angle images. Indeed, the network weights each pixel and its corresponding lens distortion prior equally regardless of the severity of distortion of the wide angle image, which affects the ability to generalize to a variety of lens distortion. 
Hence, \cite{omni} shows generalization to lens distortion closer to training distortion, and \cite{kumar2023unrectdepthnet} does not show any generalization results. 

\paragraph*{Vision transformer based approaches.} Vision transformers (ViT)~\cite{dosovitskiy2020vit} use self-attention mechanisms~\cite{vaswani2017attention} computed on image patches rather than performing convolutions. Unlike CNNs, a ViT does not have a fixed geometric structure in its architecture: any extra structure is given via positional encoding. More recently, the Swin transformer architecture~\cite{liu2021swin} incorporates a multi-scale strategy with window-based attention. Later, the Deformable Attention Transformer (DAT)~\cite{xia2022vision} adopts the concept of deformable CNNs~\cite{dai2017deformable,zhu2019deformable} to enhance transformer adaptability. \cite{zhang2023domain} proposes a distortion-aware architecture using a transformer network, but the network is limited to a fixed equirectangular distortion. Recently, methods very similar to our work, such as \cite{healswin, athwale2023darswin} use the Swin transformer \cite{liu2021swin} as their base network. 
On one hand, \cite{healswin} reasons on wide angle images by assuming a spherical projection model (\cref{sec:background}) and using a Healpix grid on sphere, unlike the Cartesian grid in the original Swin transformer. However, it does not offer generalization capabilities to unseen lenses at test time. DarSwin \cite{athwale2023darswin}, on the other hand, uses radial patches instead of the Cartesian grid in the Swin transformer network. It embeds the lens distortion parameter into the network (see \cref{sec:background} for more details), to generalize the model's performance on unseen lens distortion at test time. However, DarSwin conducts its experiments only on the image classification task. In contrast, our proposed \thename, extends \cite{athwale2023darswin} to an encoder-decoder architecture to perform pixel-level tasks, making it more effective for real-world applications.


%% file: sections/background.tex
\section{Background: Distortion-aware Radial Swin transformer (DarSwin)}
\label{sec:background}

\begin{figure}[t]
\footnotesize
\centering
\begin{tabular}{cc}
  \includegraphics[width=0.45\linewidth]{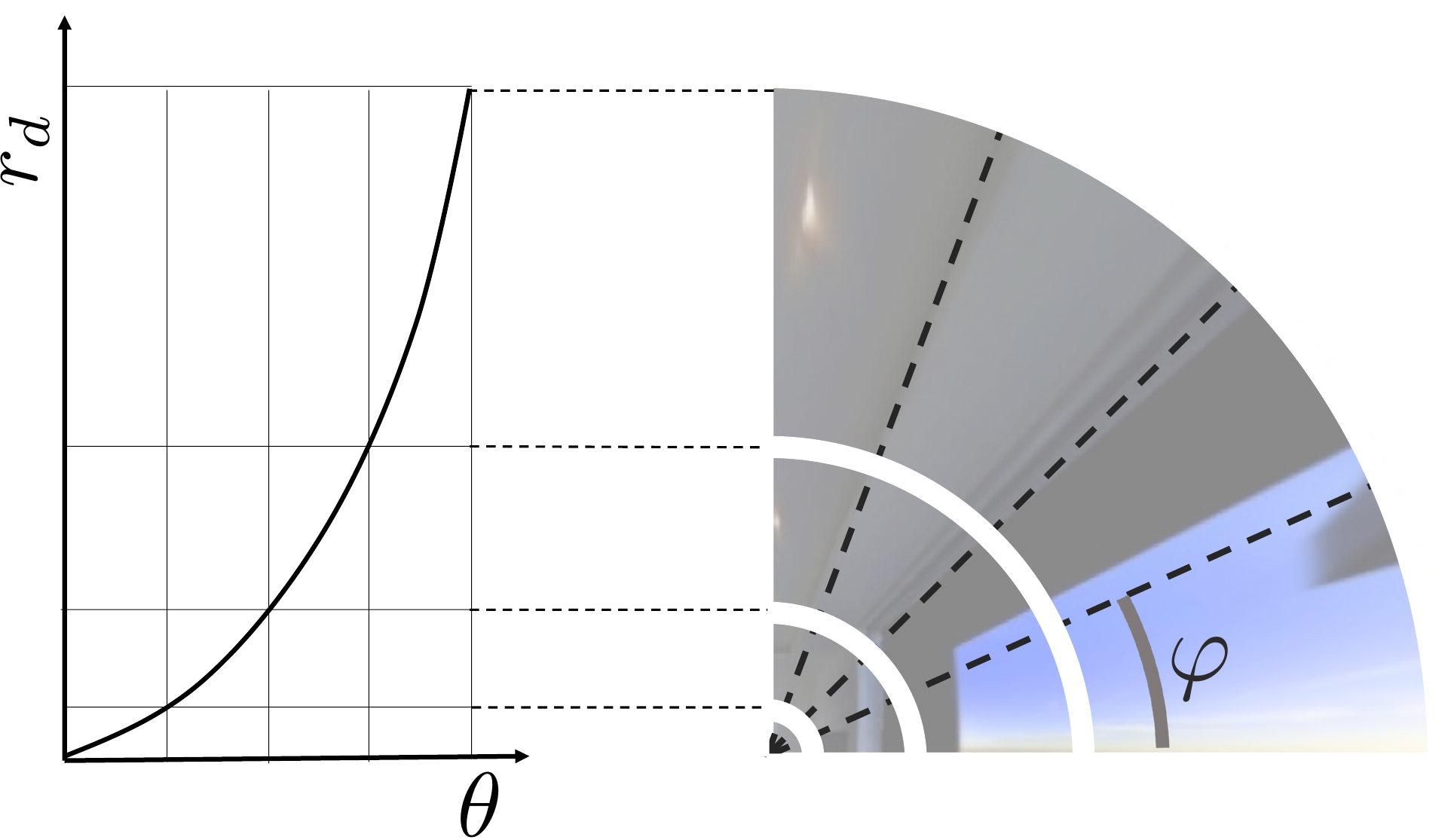} &
  \includegraphics[width=0.45\linewidth]{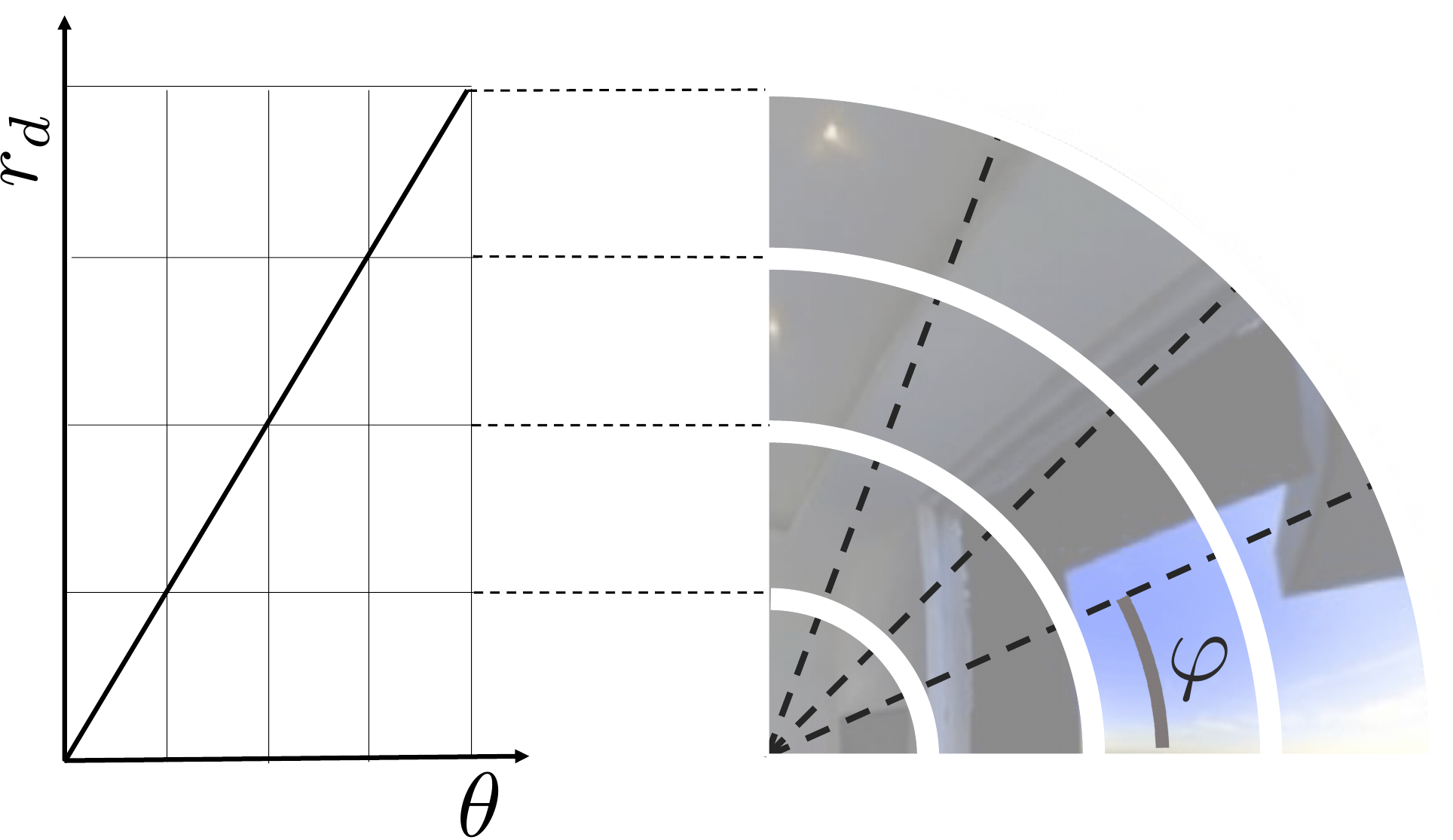} \\
  Low distortion & 
  High distortion
\end{tabular}
\caption{Radial divisions adapt to the lens distortion; here, we show low (left) and high (right) distortion for illustration purposes. DarSwin~\cite{athwale2023darswin} separates radial patches equally along $\theta$ and determines the corresponding radius on the image plane according to the (known) lens distortion curve $r_d = \mathcal{P}(\theta)$.}
\label{fig:rad}
\end{figure}

This section briefly summarizes DarSwin~\cite{athwale2023darswin}, a distortion-aware radial patch-based encoder built on the Swin Transformer~\cite{liu2021swin}. DarSwin adapts to wide angle distortions by dividing images into radial patches based on the lens distortion profile, as shown in \cref{fig:rad}.

\paragraph{Architecture overview.}

\begin{figure}[!th]
  \centering
  \includegraphics[width=1.0\linewidth]{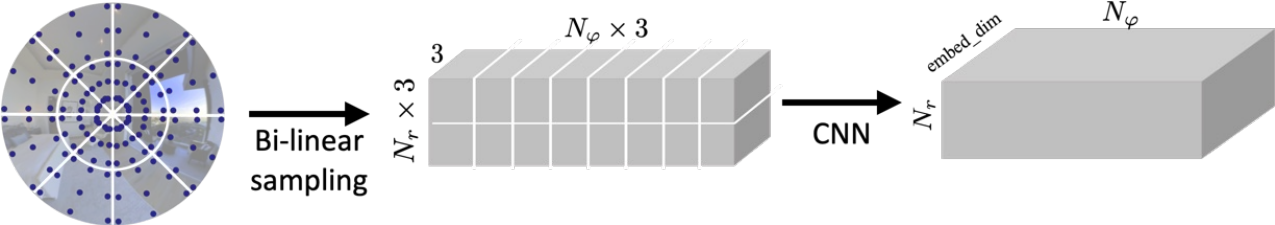}
  \caption{For illustration, the wide angle image is divided into 16 patches ($N_r = 2$ and $N_\varphi = 8$) along radius and azimuth. Nine samples are defined per patch: 3 samples along the radius and 3 samples along the azimuth. The image is bilinearly sampled and arranged in radial-azimuth format. This feature map is passed through CNN to embed each patch to get a feature map of dimension $N_r \times N_\varphi \times \text{embed-dim}$.}
  \label{fig:linear}
\end{figure}

The first layer of DarSwin divides the image into radial patches by defining the number of samples along radius $N_r$ and azimuth $N_\varphi$. Samples along azimuth are obtained directly by dividing the angular dimension of the polar representation of the image into $N_\varphi$ equal partitions as shown in \cref{fig:linear}. Samples along the radius are obtained according to the lens curve ($r_d = \mathcal{P}(\theta)$) after dividing the axis along the incident angle $\theta$ into $N_r$ equal partitions and sampling the radial value from the curve, as shown in \cref{fig:rad}. 
Moreover, the examples in \cref{fig:rad} show that this partitioning strategy allows DarSwin to adapt to any lens, knowing its distortion curve, by changing the patch size. 
A CNN is then used to linearly embed these patches. However, since the patch sizes are different and the input to the CNN must have the same dimension, a fixed set of points are sampled for each patch as shown in \cref{fig:linear}. After linear embedding, tokens are arranged in the polar format $N_r \times N_\varphi$, and are fed into the DarSwin self-attention blocks, which perform window-based self-attention. A set of non-overlapping shifted windows is defined using the patches along the azimuth dimension ($N_\varphi$), while shifts are obtained by displacing the windows along the azimuth.
Finally, a downsampling step is used to reduce the spatial resolution by merging four angular patches along the azimuth prior to the next self-attention block. It is worth noting that DarSwin uses an angular relative positional encoding technique to capture the relative information between the produced radial tokens in its attention layers.

The original paper \cite{athwale2023darswin} evaluated DarSwin on synthetically distorted ImageNet (using the Unified camera model) and demostrated its ability to adapt to new distortion curves in a zero-shot test setting.

\begin{figure*}[!ht]
\centering
\includegraphics[width=1.02\linewidth]{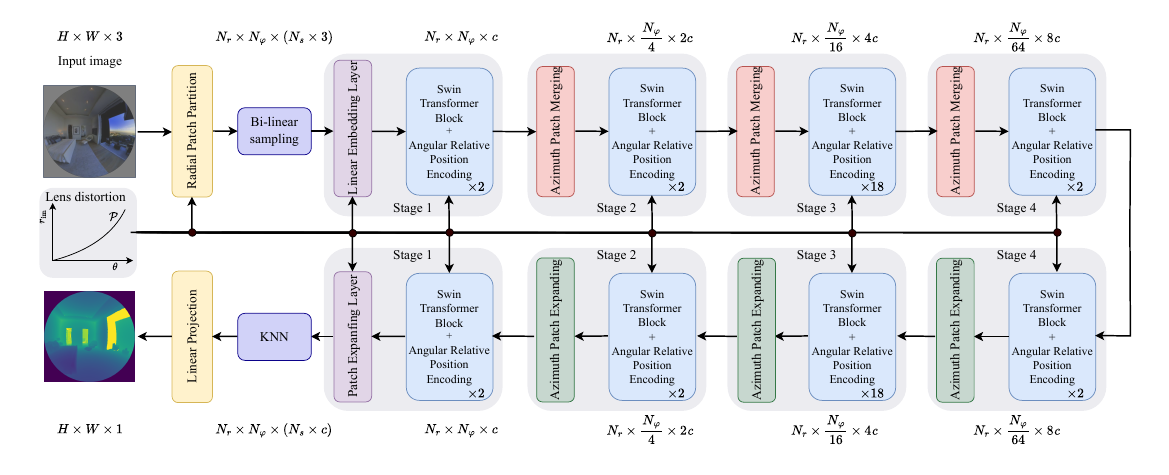}
\caption[]{Overview of our distortion-aware transformer encoder-decoder architecture, \thename. It employs hierarchical layers of DarSwin transformer blocks~\cite{athwale2023darswin} (top row) and replicates the structure in the decoder (similar to Swin-Unet~\cite{cao2021swinunet}). To make the architecture adapt to lens distortion, the patch partition, linear embedding, patch merging, and patch expanding layers, all take the lens projection curve $\mathcal{P}(\theta)$ (c.f. \cref{sec:background}) as input. The $k$-NN layer is used to project the feature map from polar ($N_r \times N_\varphi)$ to cartesian space $H \times W$. }
\label{fig:darswin-arch}
\end{figure*}


\paragraph{Unified camera model.} The Unified camera model~\cite{barreto2006unifying,mei2007single} describes the radial distortion by a \emph{single}, \emph{bounded} parameter $\xi \in [0, 1]$\footnote{$\xi$ can be slightly greater than 1 for certain types of catadioptric cameras~\cite{ying2004consider} but this is ignored here.}. It projects the world point to the image as follows
\begin{equation}
r_d = \mathcal{P}(\theta) = \frac{f\cos\theta}{\xi + \sin{\theta}} \,,
\label{eqn:sphproj}
\end{equation}
focal length, and $\xi$ the distortion parameter. We use this model for its flexibility in generating diverse distortion profiles with a single parameter $\xi$ (\cref{fig:distortion}) and its analytically invertible mapping, though our approach is not restricted to this projection model.

%% file: sections/method.tex
\section{Methodology}
\label{sec:method}


Inspired by the original DarSwin work~\cite{athwale2023darswin}, we propose two main contributions. First, we extend the DarSwin encoder-only architecture to a full encoder-decoder architecture (\cref{sec:encoder-decoder}). Second, we observe that sampling along the incident angle $\theta$ yields suboptimal coverage, leading to sparsity issues that affect performance. 
Therefore, we design a novel sampling technique (\cref{sec:sampling}) aimed at reducing sparsity by minimizing the distance between samples, thus improving performance.


\subsection{\thename architecture}
\label{sec:encoder-decoder}

\Cref{fig:darswin-arch} shows an overview of the architecture, which takes in a wide angle image and its distortion curve $\mathcal{P}(\theta)$ as input. We propose a UNet architecture for our pixel-level model, where the encoder part is a DarSwin architecture, and a decoder part incorporating two novel components: an azimuth patch expanding layer for upsampling, and a $k$-NN layer to project the outputs into the Cartesian coordinates, providing pixel-level values as explained below.


\paragraph{Azimuth patch expanding layer.}

As explained in \cref{sec:background}, DarSwin uses an azimuth patch merging layer to downsample the feature map. The radial nature of DarSwin enables various possibilities when merging patches: merging along the radius, along the azimuth, or both. Here, the encoder employs azimuth patch merging, as it is found to perform best according to \cite{athwale2023darswin}. Consequently, we propose an azimuth patch expanding strategy.


As in \cite{cao2021swinunet}, we use an MLP for the expanding layer. 
We use this layer along the azimuth dimension to upsample by a $4 \times$ factor. For example, consider the first (rightmost) patch expanding layer in \cref{fig:darswin-arch}. The input feature map $(N_r \times \frac{N_\varphi}{64} \times 8c)$ is first given to an MLP layer to expand the feature dimension by $4 \times$ to get $(N_r \times \frac{N_\varphi}{64} \times 32c$) where $N_r \text{ and } N_\varphi$ are number of divisions along radius and azimuth respectively (c.f. \cref{sec:background}). The feature map is then rearranged to reduce the feature dimension and increase the resolution of the feature map along the azimuth dimension to obtain $(N_r \times \frac{N_\varphi}{16} \times 4c$). 

\paragraph{$k$-NN layer.}

Lastly, we employ a $k$-NN layer to map the polar feature map back to Cartesian coordinates. Each pixel coordinate in the image is associated with its $k$ closest samples (we use $k=4$), and their respective feature vectors are averaged. Since sample point locations are known, the $k$-NN layer is fixed and not trainable. The $k$-NN output (of dimensions $H\times W\times c$) is fed into the last linear projection layer to get the desired output for the required task.

\subsection{Proposed sampling method}
\label{sec:sampling}

\begin{figure}[t]
\footnotesize
\centering
\setlength{\tabcolsep}{1pt}
\begin{tabular}{cccc}
\rotatebox{90}{\hspace{3em} $\xi = 0$} & 
\includegraphics[width = 0.3\linewidth]{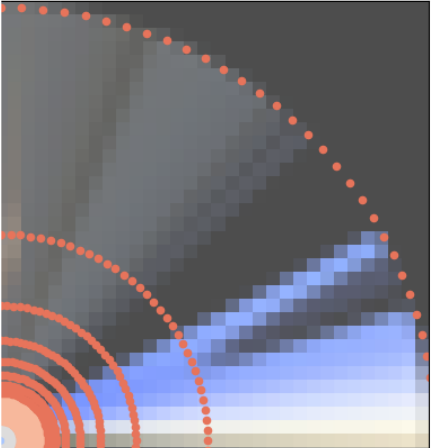} &
\includegraphics[width = 0.3\linewidth]{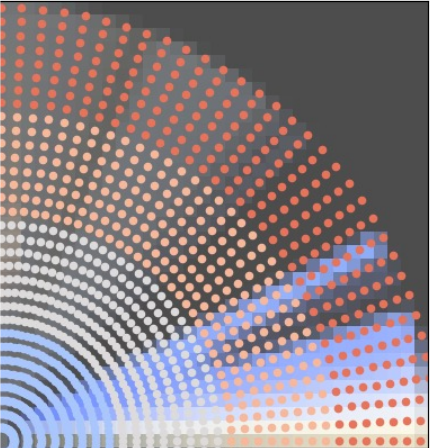} &
\includegraphics[width = 0.3\linewidth]{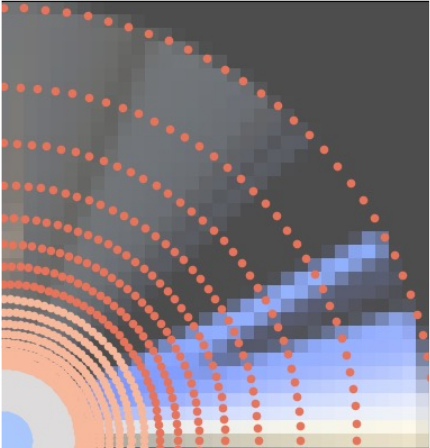} \\
\rotatebox{90}{\hspace{3em} $\xi = 1$} & 
\includegraphics[width = 0.3\linewidth]{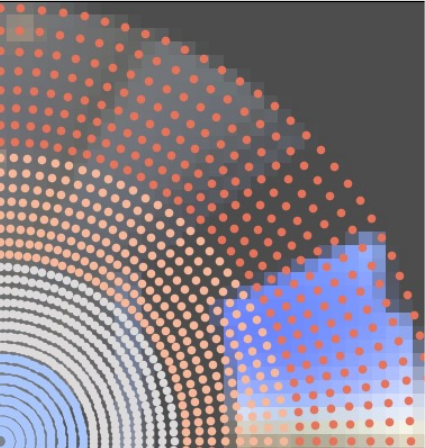} &
\includegraphics[width = 0.3\linewidth]{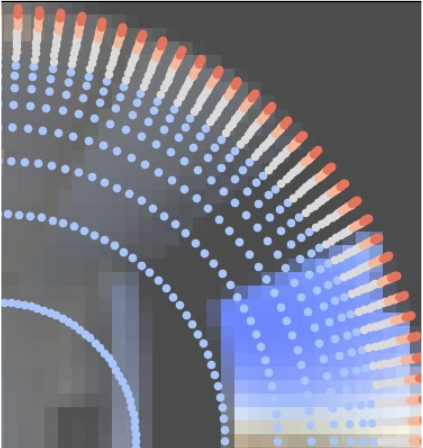} &
\includegraphics[width = 0.3\linewidth]{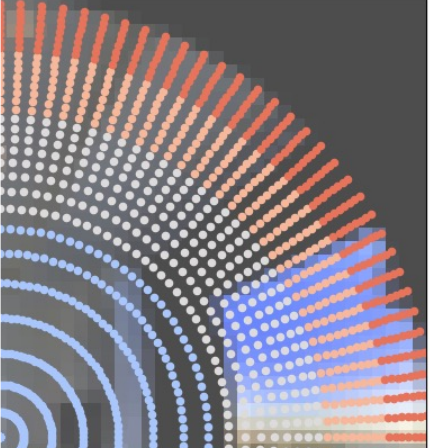} \\
& 
(a) $r_d = \mathcal{P}(\theta)$ & 
(b) $r_d = \mathcal{P}(\tan \theta)$ & 
(c) $r_d = \mathcal{P}(g(\theta))$ 
\end{tabular}
\caption{Illustration of sampling (represented by colored dots) on a quadrant of an image taken from two different lenses ($\xi$ = 0 (top row) and $\xi = 1$ (bottom row). The images is sampled according to the lens distortion curve $\mathcal{P}$ applied on different functions of $\theta$: (a) $\theta$, (b) $\tan \theta$, and (c) our novel $g(\theta)$. Observe how the first two options create large holes at either extreme values of $\xi$. In contrast, our proposed function offers a good compromise across a wide range of distortions.}
\label{fig:sparse}
\end{figure}
\begin{figure}[t]
\centering
\footnotesize
\includegraphics[width=1.0\linewidth]{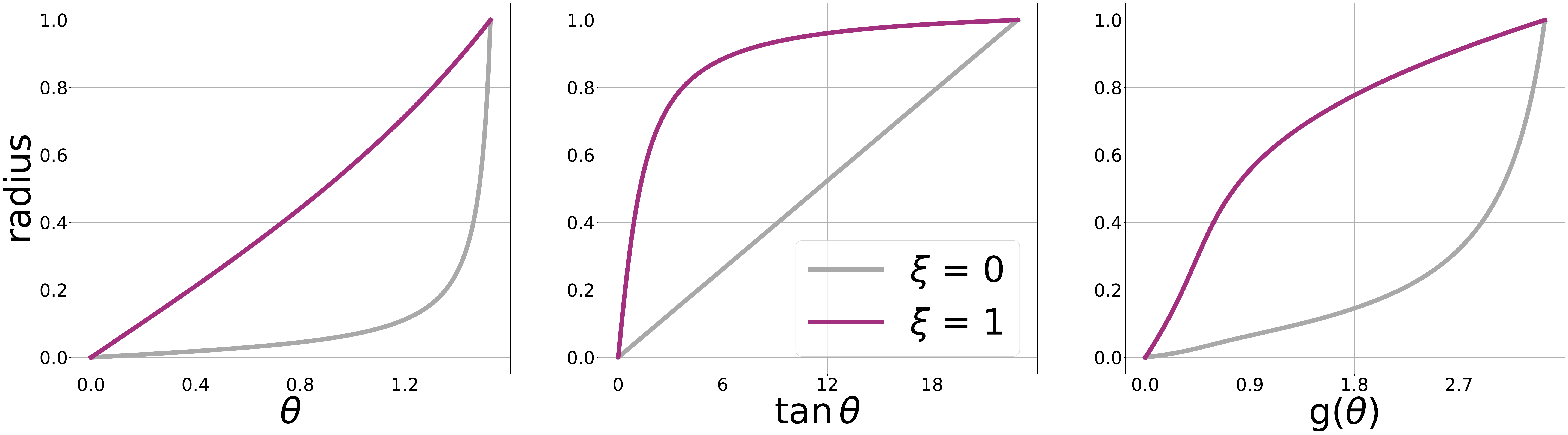}
\caption{Lens distortion curves for least ($\xi=0$) to most ($\xi=1$) distorted using the unified camera model for illustration. We represent the same curves according to, from left to right, $\tan \theta$, our new $g(\theta)$, and $\theta$. The high slopes present in both $\tan \theta$ and $\theta$ curves mean that samples will be spread far apart on the image plane. In contrast, our $g(\theta)$ offers a good compromise across the range of distortions.}
  \label{fig:curve}
\end{figure}

In the original DarSwin architecture, the input to the linear embedding layer (\cref{fig:darswin-arch}) must have the same dimension. Therefore, a fixed set of points are sampled from the image for each patch. However, because the patch dimensions change according to the distortion, this can create sparse samples along the radius. \Cref{fig:sparse} shows the samples obtained with two extreme distortions using the unified camera model with $\xi = 0$ and $\xi = 1$, respectively. \Cref{fig:sparse}-(a) shows how sampling according to the lens curve, $\mathcal{P}(\theta)$, used in \cite{athwale2023darswin}, results in significant sparsity in the image, where many samples are missing across the radius, under the perspective projection ($\xi=0$). Sampling according to another function of $\theta$, for example $\mathcal{P}(\tan \theta)$ in \cref{fig:sparse}-(b), enhances the results but creates sparse samples when the distortion is very high (e.g., $\xi=1$). Here, we are looking for another function of $\theta$, named $g(
\theta)$ which spreads out samples as uniformly as possible across a wide range of distortions (\cref{fig:sparse}-(c)). The lens functions are plotted in \cref{fig:curve}. We observe that when the derivative (slope) of the lens function is high, the samples are spread far apart in the image. 

More formally, we are looking for a strictly monotonic function $g$ that has minimal derivative over its entire range when applied under both low and high distortion. As a proxy for representing distortion, we again employ the unified camera model (c.f. \cref{sec:background}) and are looking for a function $g$ that minimizes
\begin{equation}
\max_\theta \left(\frac{d \mathcal{P}(g(\theta))|_{\xi = 0}}{d (g(\theta))}\right)
+ 
\max_\theta \left(\frac{d \mathcal{P}(g(\theta))|_{\xi = 1}}{d (g(\theta))}\right) \,.
\label{equn:metric}
\end{equation}

We minimize the derivative only for low and high distortions, as the derivatives for all intermediate distortions lie between these two extremes (details in the supplementary). The function $g$ is parameterized as a convex combination of two monotonic functions $p_n$ and $q_m$.
\begin{align}
    &g(\theta) = \lambda p_n(\theta) + (1-\lambda)q_m(\theta)\,, \\ 
    \text{with } 
    &p_n(\theta) = b\left(\frac{\theta}{a}\right)^n \text{ and  }
    q_m(\theta) = 1 - \left( 1 - \frac{\theta}{a}\right)^m \,. \nonumber 
\end{align}
This ensures that the resulting curve is monotonic (see the supplementary for more details). We search the optimal parameters $\lambda$, $m$, $n$ and $b$, that minimize the objective in \cref{equn:metric}. For optimization, we perform an exhaustive search for $\lambda \in [0, 1]$ with 10 steps, $m \in [1, 20]$ with 60 steps, $n \in [0.5, 5]$ with 20 steps, and $b \in [2, 10]$ with 40 steps. We find that the values $\lambda =0.777, m = 5.5084, n =5.0, a = \frac{\text{FOV}}{2}, b=4.1052$ gives us an optimal curves for both $\xi = 0$ and $\xi = 1$ as shown in \cref{fig:curve} (right). Using this optimal curve $g(\theta)$ for sampling reduces sparsity at both extreme cases (i.e., zero and maximal distortion levels) compared to the previous methods (\cref{fig:sparse}-(c)). In our experiments we sample 25 points along azimuth and 4 points along radius, in total 100 sample points per patch. We also analyze the performance on the depth estimation task using these three functions in supplementary material. 

%% file: sections/experiments.tex
\section{Depth estimation experiments}
\label{sec:depth}

To evaluate the efficacy of \thename's generalization and robustness on unseen distortion profiles, we perform monocular depth estimation experiments using synthetically generated wide angle images using a panoramic dataset \cite{Matterport3D}.

\begin{figure}[t]
\centering
\includegraphics[width=0.8\linewidth, angle=0]{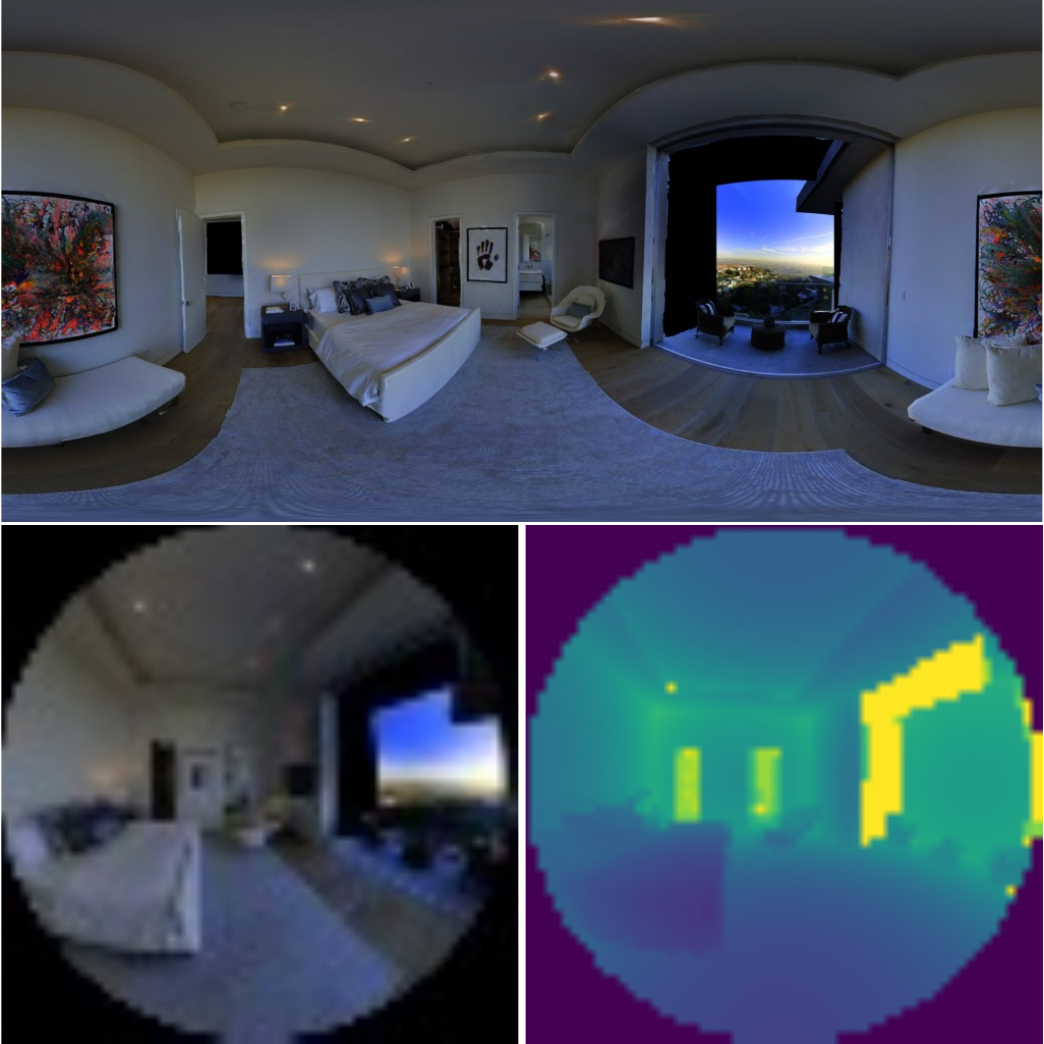}
\caption{From a $360^\circ$ panorama (top) from the Matterport3D dataset~\cite{Matterport3D}, we generate a wide angle image and its depth map (bottom) with a field of view $175^\circ$. Lens distortion is simulated with the uniform camera model (here, $\xi = 0.95$).}
\label{fig:dataset_pixel}
\end{figure}

\subsection{Datasets}
\label{sec:dataset_gen}

Existing wide-angle depth estimation datasets, such as Woodscapes~\cite{yogamani2019woodscapes}, lack the diverse distortion profiles required to evaluate our network's generalization. To address this, we generate synthetic wide-angle images by cropping $175^\circ$ field-of-view images from Matterport3D panoramas~\cite{Matterport3D} and simulating lens distortion using the unified camera model~\cite{barreto2006unifying, mei2007single} (see \cref{sec:background}), as illustrated in \cref{fig:dataset_pixel}.

\begin{figure}[t]
\footnotesize
\setlength{\tabcolsep}{1pt}
\begin{tabular}{cccc}
Very low & Low & Medium & High \\
\includegraphics[width = 0.24\linewidth]{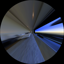} &
\includegraphics[width = 0.24\linewidth]{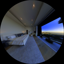} &
\includegraphics[width = 0.24\linewidth]{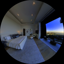} &
\includegraphics[width = 0.24\linewidth]{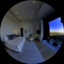} 
\end{tabular}
\caption{Visualization of a wide angle crop from a panorama with different distortions representing 4 different distortion levels From left to right: very low, low, medium, and high, used in four different training sets as explained below. The image is cropped from panorama with an original resolution of $512 \times 1024$, where the generated wide angle image is subsequently down-sampled to $64 \times 64$ after warping.}
\label{fig:distortion}
\end{figure}

\paragraph*{Training set.}

Similar to DarSwin \cite{athwale2023darswin}, we generate four different training sets with different levels of distortion, defined by the distortion parameter $\xi$: ``very low'' ($\xi \in [0.0, 0.05]$), ``low'' ($\xi \in [0.2, 0.35]$), ``medium'' ($\xi \in [0.5, 0.7]$), and ``high'' ($\xi \in [0.85, 1.0]$) as illustrated in \cref{fig:distortion}.

Training images are synthetically generated from panoramas on the fly during training with distortion $\xi$ sampled from their respective intervals, and the yaw viewing angle in the panorama is uniformly sampled in the $[0, 360^\circ]$ interval. Each of the four training sets (one for each distortion group) contains $9,180$ panoramas of original resolution of $512 \times 1024$, where the generated wide angle image is subsequently downsampled to $64 \times 64$ after warping.

\paragraph*{Test set.}

To evaluate performance and zero-shot generalization to seen and unseen distortion profiles, we generate 20 test sets, each with a fixed distortion value uniformly sampled from $\xi \in [0, 1] $ in steps of 0.05. All test sets are created from the same $1,620$ test panoramas.

\begin{figure}[!t]
\centering
\footnotesize
\begin{tabular}{cc}
    \includegraphics[width=0.45\linewidth]{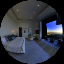} & 
    \includegraphics[width=0.45\linewidth]{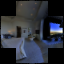} \\
    (a) Wide angle image & 
    (b) Piecewise perspective
\end{tabular}
\caption{(a) $175^\circ$ field of view wide angle image and (b) corresponding undistorted image using piecewise linear correction (cubemap representation).}
\label{fig:cubemap}
\end{figure}

\subsection{Baselines}

We compare to the following baselines. First, we use Swin-Unet~\cite{cao2021swinunet} and Swin-UPerNet (a Swin~\cite{liu2021swin} encoder with UPerNet~\cite{xiao2018unified} decoder). We also compare with DAT~\cite{xia2022vision}, which leverages deformable attention in order to understand lens distortion for better robustness. Hence, we compare with DAT-UPerNet (a DAT~\cite{xia2022vision} encoder with UPerNet decoder). Since these baselines do not have access to the lens distortion, as opposed to our proposed \thename, we also correct the distortion in the image and train the Swin baselines on this input. We dub these alternatives as Swin-Unet(undis) and Swin-UPerNet(undis). 

\paragraph*{Undistorting with piecewise perspective projection.}
Undistorting a $175^\circ$ field-of-view wide angle image to a single perspective image will result in extremely severe stretching. Instead, we follow the piecewise perspective correction strategy in \cite{yogamani2019woodscapes} and undistort the image to a partial cubemap, which is composed of 6 perspective faces of $90^\circ$ field of view each, unrolled into an image and cropped to keep only the valid pixels. As shown in \cref{fig:cubemap}, this preserves the entire field of view while minimizing stretching. As mentioned above, these images are used to train the Swin-Unet(undis) and Swin-UPerNet(undis) baselines.

\begin{figure*}[!ht]
    \centering
    \includegraphics[width=1.0\linewidth]{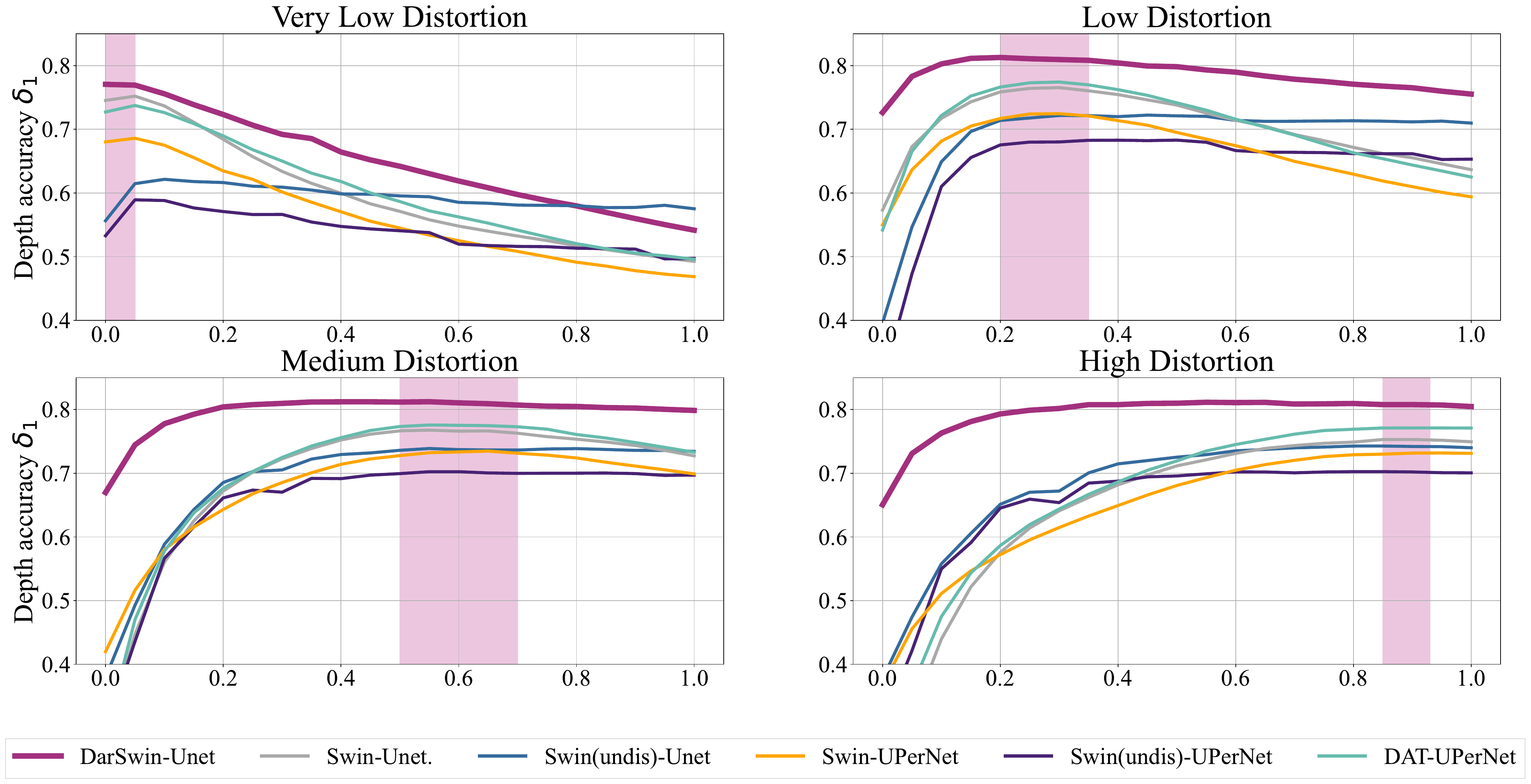}
    \caption[]{Depth estimation accuracy $\delta_1$ (higher is better) as a function of test distortion for: \thename, Swin-Unet \cite{cao2021swinunet}, Swin(undis)-Unet, Swin-UPerNet \cite{liu2021swin, xiao2018unified}, Swin(undis)-UPerNet and DAT-UPerNet \cite{xia2022vision, xiao2018unified}. All methods are trained on a restricted set of lens distortion curves (indicated by the pink shaded regions): (a) Very low, (b) low, (c) medium, and (d) high distortion. We study the generalization abilities of each model by testing across $\xi \in [0, 1]$. We can see that the performance of all the baseline decreases as we move away from training distortion, but the curve for \thename remains relatively flat, indicating that \thename can generalize on unseen wide angle distortion at test time.}
    \label{fig:results-depth}
\end{figure*}

\subsection{Training details and evaluation metrics}

All baselines have 1024 patch divisions on image size $64 \times 64$ with patch size $2 \times 2$ and window size  $4 \times 4$ along the height and width.
For \thename, we employ 16 divisions along the radius and 64 on the azimuth on an image with a total of 1024 divisions (to have the same number of patches as baselines).
All encoders (Swin and DAT) are first pre-trained for classification on the distorted tiny-ImageNet dataset from \cite{athwale2023darswin}. Pre-trained encoders, along with their respective decoders, are fine-tuned on the depth estimation task.
All methods are trained with the SGD optimizer with momentum $0.9$ and weight decay $10^{-4}$ with a batch size of $8$. We employ a polynomial learning rate policy with a base learning rate of $0.01$ and $\text{power}=0.9$. We use random flips and rotations as data augmentation. 

\paragraph*{Training loss.}
Since the global scale of a scene is a fundamental ambiguity in depth estimation \cite{eigen2014depth}, we train the network using the scale-invariant loss in log space \cite{shu2022sidert}:
\begin{equation}
    \ell = \sqrt{\frac{1}{n} \sum_{i}^{} {d_i}^2 - \frac{\lambda}{n^2} \left(\sum_{i}^{} {d_i}\right)^2} \,,
\end{equation}
where $d_i$ is the difference between the predicted and ground truth (log-)depth. We use $\lambda=0.85$.

\paragraph*{Evaluation metrics.}
\label{sec:depth_metrics}

We evaluate performance on typical depth estimation metrics~\cite{shu2022sidert}: absolute relative error, RMSE, log-RMSE, squared relative error, and accuracy under threshold ($\delta_i, \, i \in \{1,2,3\}$). The paper reports results on 
\begin{align}
    \delta_1 = \frac{1}{|D|} |\{ d\in D | \max(\frac{d^*}{d},\frac{d}{d^*}) \leq 1.25\} \,,
\end{align}
where $D$, $d^*$, and $d$ are the set of valid, ground truth and predicted depths, respectively. Please consult the supplementary material for results on  other metrics.

\subsection{Zero-shot generalization}

We perform a similar generalization test as \cite{athwale2023darswin}, we train all the baselines and \thename on all four training sets with different levels of distortion independently (represented by the pink shaded region in \cref{fig:results-depth}, and evaluate them on all of the 20 test sets, as explained above. Our primary focus is on the efficacy of the network on unseen lens distortion (outside the pink shaded region). As shown in \cref{fig:results-depth}, we can see that for all the methods including \thename the depth estimation accuracy $\delta_1$ metric is highest in the pink shaded region for each training set since the model has seen those lens distortion while training. But as we move away from the pink region, the performance for the baselines decreases rapidly, as these lens distortions are not present during test time, but \thename maintains its performance even outside the training distortion region for ("low",  "medium" and "high" distortion training sets). When \thename is trained in "very low" distortion, we see a decrease in performance as we move away from training distortion, but still \thename outperforms all the baselines.

\thename demonstrates better generalization capabilities across different lenses by embedding the distortion parameter within the network, as introduced by DarSwin \cite{athwale2023darswin}. The change in patch size, as depicted in \cref{fig:rad}, allows each patch in the attention layer to be weighted based on the specific lens distortion.

For a fair comparison, the baselines Swin-Unet(undis) and Swin-UPerNet(undis) are equiped with the distortion parameter knowledge as well. However, despite this inclusion, these baselines fail to generalize effectively to other lenses. The primary reason for this is the presence of artifacts resulting from the undistortion process.

\subsection{Ablations}  

\paragraph{$k$-NN analysis.}

\begin{figure}[!t]
\footnotesize
\setlength{\tabcolsep}{1pt}
\begin{tabular}{ccccc}
Ground truth  & \multicolumn{4}{c}{
Reconstructed labels by $k$-NN}   \\
 $\xi = 0.25$& $\hspace{12pt} 4\times4$& $8\times4$ & $16\times4$ & $25\times4$ \\
\includegraphics[width = 0.18\linewidth]{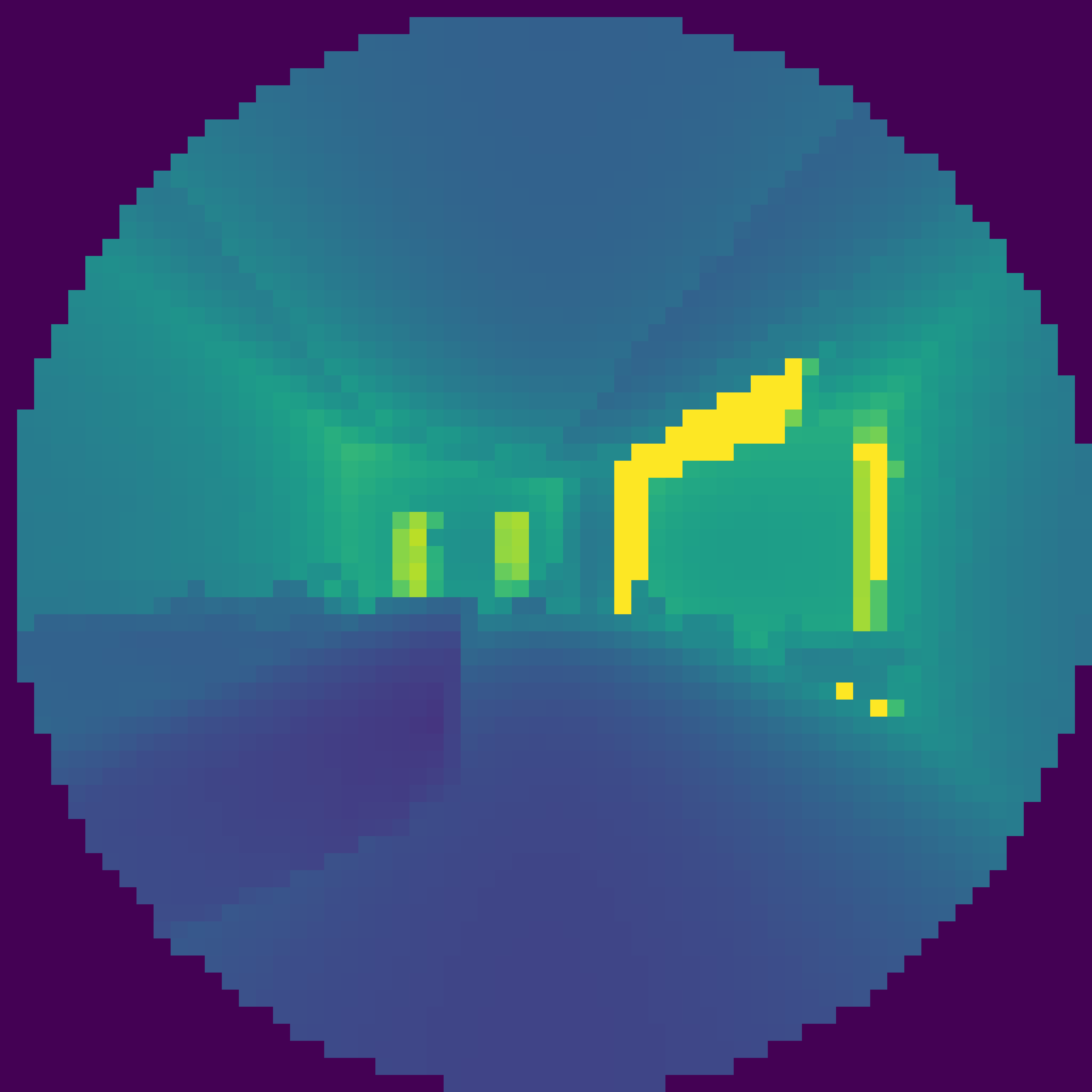}&
\hspace{10pt}
\includegraphics[width = 0.18\linewidth]{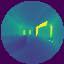} &
\includegraphics[width = 0.18\linewidth]{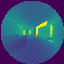} 
& \includegraphics[width = 0.18\linewidth]
{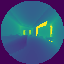}&
\includegraphics[width = 0.18\linewidth]{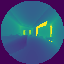}
\end{tabular}
\caption{Visualizations of reconstructed labels by the $k$-NN layer considering different numbers of samples per patch. For a low number of samples ($4\times4$ per patch), we notice some artifacts, particularly in the center. These artifacts progressively disappear when increasing the number of samples, leading to faithful reconstructions with the $k$-NN $25\times4$.}
\label{fig:knn_ab}
\end{figure}

We study the impact of the number of samples per patch on the reconstruction quality. As explained in \cref{sec:encoder-decoder}, 25 points are sampled along the radius, and 4 points are sampled along the azimuth, giving $25 \times 4$ samples for each patch. 
Since the $k$-NN layer is not trainable, ablations on this part of the model are made based on the ground truth labels. We distort depth maps from Matterport3D \cite{Matterport3D} with $\xi = 0.25$. The ground truth label is sampled and reconstructed using the sample locations and the $k$-NN layer.
To evaluate the efficiency of this layer, we calculate the Mean Absolute Error (MAE) over valid pixels between each ground truth label and its corresponding reconstructed label, and then we average values over all images. We ablate on the number of samples per patch as shown in \cref{tab:ablation_knn} and illustrated in \cref{fig:knn_ab}. We show that $25\times4$ samples per patch results in efficient projection from polar features to cartesian features with an error of 0.8\% and 10.3ms/image compared to 3.1ms/image for $4\times4$ samples.

\begin{table}[!ht]
\footnotesize
\centering
\caption{Ablation study on the efficiency of the $k$-NN layer: for a limited number of samples per patch ($2\times2$ and $4\times4$), we have an important error of 21.3\% and 9.3\%, respectively. The error reduced significantly for a number of samples equal or higher to $8\times8$.}
\begin{tabular}{lcccc} 
\toprule
\# samples/patch   & $4\times4$ & $8\times4$  & $16\times4$ & $25\times4$  \\ 
\midrule
MAE       & 4.09\% &  2.36\%   & 1.28\% &  0.8\%\\
\bottomrule
\end{tabular}
\label{tab:ablation_knn}
\end{table}

\begin{figure}
    \centering
    \includegraphics[width=0.9\linewidth]{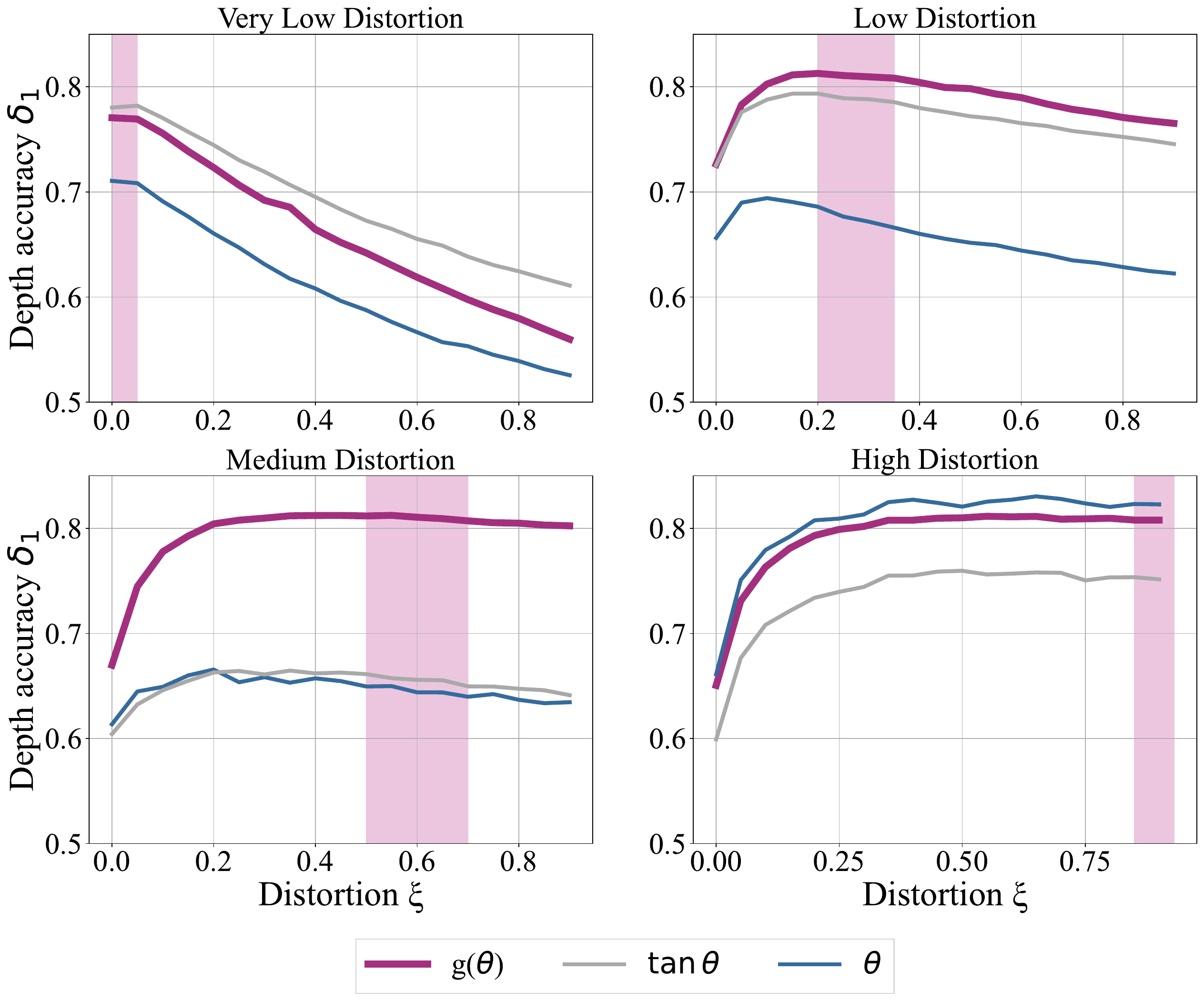}
    \caption{Depth accuracy when the model is trained across all four distortion levels—"very low," "low," "medium," and "high"—using different sampling strategies. Our proposed sampling strategy, $\mathcal{P}(g(\theta))$, demonstrates better performance in generalization across all distortion levels compared to $\mathcal{P}(\theta)$ and $\mathcal{P}(\tan\theta)$.}
    \label{fig:fig_sample}
\end{figure}

\paragraph{Sampling function.} We compared the choice of sampling strategy (\cref{sec:sampling}), using depth accuracy for the model trained on each training set and tested on all $\xi$ $\in [0, 1]$. We experiment on the three distortion curves $\text{radius} \text{ vs }(\tan \theta, \theta, g(\theta))$. 
The generalization performance of curves with respect to $\tan(\theta)$ and $\theta$ surpasses our method $g(\theta)$ in specific cases—"very low" and "high" distortion (see \cref{fig:fig_sample}). This behavior aligns with \cref{fig:sparse}, as $\tan(\theta)$ benefits from dense sampling near $\xi = 0$, making it effective at low distortion but less so elsewhere. Similarly, $\theta$ performs well at high distortion due to dense sampling near $\xi = 1$. However, our proposed sampling method with respect to $g(\theta)$ consistently outperforms across all distortion levels.



%% file: sections/discussion.tex
\section{Discussion}
\label{sec:discussion}


This paper introduces \thename, a novel radial-based distortion-aware encoder-decoder transformer built upon DarSwin \cite{athwale2023darswin}. \thename dynamically adapts its structure according to the lens distortion profile of a calibrated lens, enabling it to achieve state-of-the-art performance in zero-shot adaptation on various lenses for the depth estimation.

A key contribution of our work is the development of a novel sampling function designed to address the sparsity issues inherent in distortion-based sampling techniques introduced by \cite{athwale2023darswin}. This improvement is particularly important for pixel-level tasks, where sparsity in sampling has a more pronounced impact and can significantly degrade performance, unlike in classification tasks.

\paragraph*{Limitations and future work} 


While \thename shows significant advancements in distortion-aware pixel-level tasks like depth estimation, it has limitations. Scaling to high-resolution images remains challenging, but incorporating ideas from SwinV2 \cite{liu2022swintransformerv2scaling}, which excels at scaling Swin Transformer architectures, could enhance its ability to handle higher resolutions. Additionally, \thename relies on prior knowledge of lens distortion profiles, limiting its use in uncalibrated scenarios. Future work could address this by integrating a secondary network to predict distortion parameters or developing an end-to-end model that learns distortion directly from input images.

%% file: sections/acknowledgment.tex
{\small 
\paragraph{Acknowledgments} This research was supported by NSERC grant ALLRP-567654, Thales, an NSERC USRA to E. Bergeron, Mitacs and the Digital Research Alliance Canada. We thank Yohan Poirier-Ginter, Frédéric Fortier-Chouinard and Justine Giroux for proofreading.}